%% file: egpaper_final.tex
\documentclass[10pt,twocolumn,letterpaper]{article}

\usepackage{cvpr}
\usepackage{times}
\usepackage{epsfig}
\usepackage{graphicx}
\usepackage{amsmath}
\usepackage{amssymb}

\usepackage{adjustbox}
\usepackage{array}
\usepackage{dsfont}
\usepackage{pifont}
\usepackage{subcaption}
\usepackage{booktabs}

\usepackage[pagebackref=true,breaklinks=true,letterpaper=true,colorlinks,bookmarks=false]{hyperref}

\cvprfinalcopy 

\def\cvprPaperID{23} 

\ifcvprfinal\fi
\begin{document}

\title{Budget-aware Semi-Supervised Semantic and Instance Segmentation}



%



\author{Miriam Bellver$^1$, Amaia Salvador$^2$,
Jordi Torrres$^1$ and Xavier Giro-i-Nieto$^2$
\\ \vspace{2mm}
$^1$Barcelona Supercomputing Center  \:\:\:
$^2$Universitat Polit\`ecnica de Catalunya}


\maketitle
\begin{abstract}
    Methods that move towards less supervised scenarios are key for image segmentation, as dense labels demand significant human intervention. Generally, the annotation burden is mitigated by labeling datasets with weaker forms of supervision, e.g. image-level labels or bounding boxes. Another option are semi-supervised settings, that commonly leverage a few strong annotations and a huge number of unlabeled/weakly-labeled data. In this paper, we revisit semi-supervised segmentation schemes and narrow down significantly the annotation budget (in terms of total labeling time of the training set) compared to previous approaches. With a very simple pipeline, we demonstrate that at low annotation budgets, semi-supervised methods outperform by a wide margin weakly-supervised ones for both semantic and instance segmentation. Our approach also outperforms previous semi-supervised works at a much reduced labeling cost. We present results for the Pascal VOC benchmark and unify weakly and semi-supervised approaches by considering the total annotation budget, thus allowing a fairer comparison between methods. 
    
    
\end{abstract}

\input{sections/1_introduction.tex}
\input{sections/2_related.tex}
\input{sections/3_benchmark.tex}

\input{sections/4_semisupervised_setup.tex}

\input{sections/5_semanticseg.tex}

\input{sections/6_instanceseg.tex}
\input{sections/7_conclusions.tex}

{\small
\bibliographystyle{ieee}
\bibliography{egpaper_final}
}

\end{document}

%% file: sections/1_introduction.tex
\section{Introduction}
\label{sec:intro}

In computer vision, current state-of-the-art models based on Convolutional Neural Networks are data-hungry, and their performance is related to the amount of annotated data available for training. In particular, segmentation annotations are very costly, as they require a label for each pixel of the image.  Therefore, there is a growing interest in training segmentation models that do not rely on a high annotation budget but still achieve a competitive performance. 

For semantic and instance segmentation, the use of weak labels as a cheaper supervision signal to train segmentation models has been extensively explored in the literature. Some of the most popular weak supervision signals are image-level labels~\cite{wei2018revisiting, zhang2018decoupled, ahn2018learning, zhou2018weakly} or bounding boxes~\cite{dai2015boxsup, papandreou2015weakly, khoreva2017simple, li2018weakly, zhao2018pseudo}. Although the results are promising, they are still far from the performance of methods that rely on stronger supervision.

\begin{figure}
  \centering
  \includegraphics[width=\columnwidth]{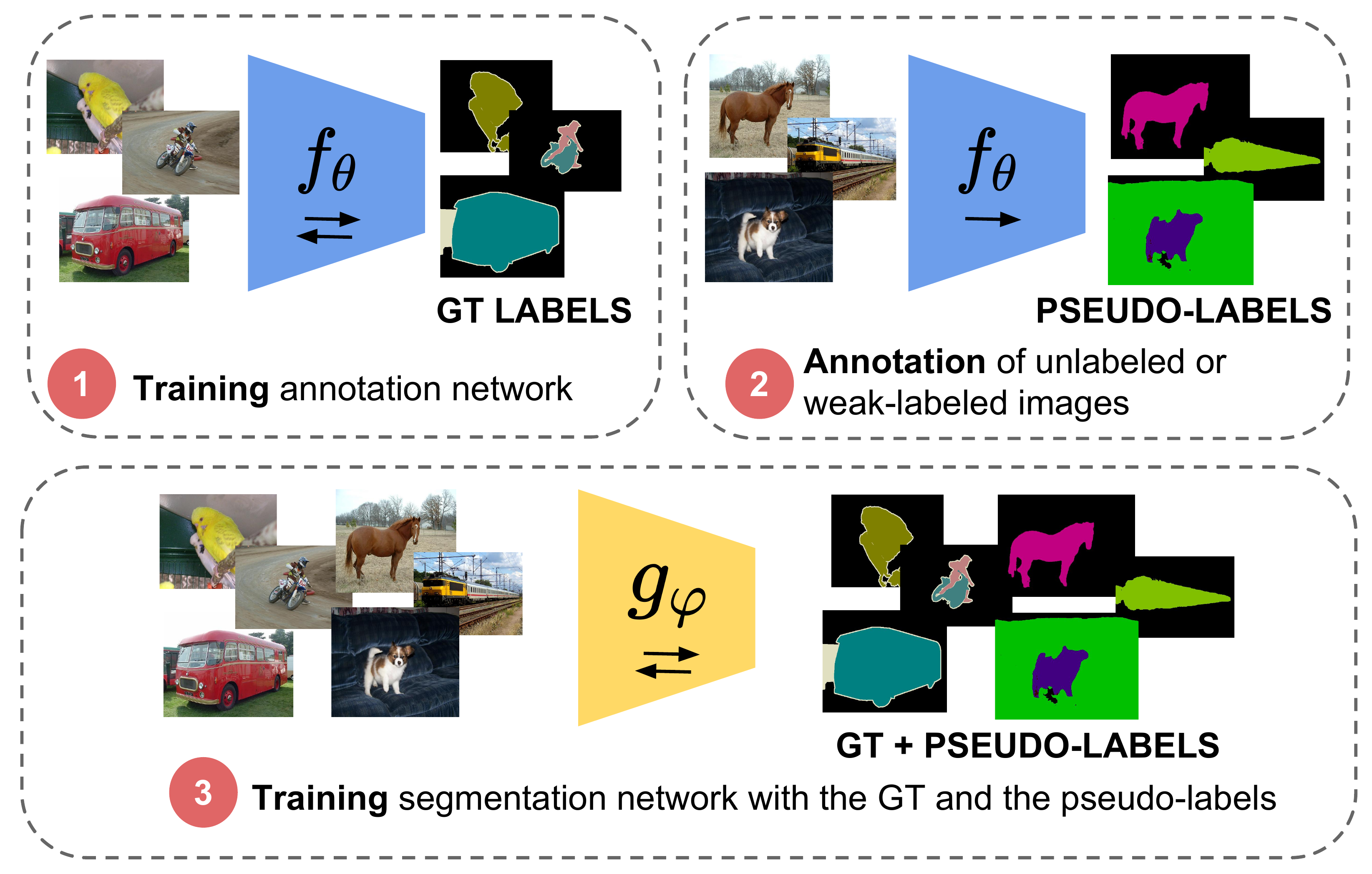}
  \caption{Our semi-supervised training pipeline consists of two networks, an annotation network trained with strong supervision, and a segmentation network trained with the union of pseudo-annotations and strong-labeled samples.}
  \label{fig:scheme}
  \vspace{-3.25mm}
\end{figure}

Another option to lower the annotation cost are semi-supervised scenarios, where a small subset of the data is strongly annotated, and the remaining samples are unlabeled/weakly-labeled. The most successful semi-supervised methods handle heterogeneous annotations (few strong and a huge amount of weak labels) and, although they reach higher performance ~\cite{papandreou2015weakly, hong2015decoupled, wei2018revisiting}, their annotation cost is much higher than the one related to weakly-supervised schemes. 

The goal of weakly and semi-supervised methods is to obtain segmentation results that are competitive with their fully-supervised counterparts, while requiring a much lower annotation cost. However, previous works do not typically compare to each other in terms of the annotation budget. In this paper, we argue that when the goal is to minimize human effort, methods should be compared considering the annotation cost regardless of the type of annotation they use. In this direction, \cite{bearman2016s} proposed a comparison between weakly and fully-supervised semantic segmentation methods that contemplates the total annotation time required for the training set. We extend this analysis including semi-supervised methods and also, for the first time, for the instance segmentation task. This will allow a unified analysis across different supervision setups (weakly- or semi-supervised) and different supervision signals, comparing the total annotation time when fixing a certain budget.

In this work, we present a semi-supervised scheme trained with low annotation budgets that reaches significantly better performance than weakly-supervised methods while having the same annotation cost. Our proposed pipeline consists of two networks: a first annotation model that generates pseudo-annotations for the unlabeled or weakly-labeled data, and a second segmentation model that is trained with both the strong and pseudo-annotations (Figure~\ref{fig:scheme}). In order to lower the annotation budget, first we 
combine strongly annotated data with unlabeled data,
so that only strong annotations have an associated annotation cost. With only a few strong annotations, we reach higher performance than previous weakly and semi-supervised approaches for both semantic and instance segmentation, at much reduced annotation budgets. 

We also analyze heterogeneous annotations for instance segmentation, which combine both strong and weak labels. The weak label that we choose consists in counting the number of objects there are for each of the class categories of the dataset~\cite{gao2017c}. To the authors knowledge, this is the first time that image level labels with object counts are used for instance segmentation. 
We propose to exploit weak labels by feeding them into the annotation network. As weak labels involve a cost, we adjust the number of samples to analyze different supervision scenarios. We find that, for low annotation budgets, this solution outperforms the standard semi-supervised pipeline.

Our contributions can be summarized as follows: 1) We unify the segmentation benchmarks regardless of the training setting and the supervision signals by comparing them in terms of the total annotation cost they require, 2) we outperform previous semi-supervised semantic segmentation methods at low annotation budgets for the Pascal VOC benchmark~\cite{everingham2010pascal}, and present the first quantitative results for semi-supervised instance segmentation for this dataset when no extra images are available, 3) we show that when fixing a low annotation budget, it is more convenient having fewer but stronger-labeled data over having larger weakly-annotated sets.


%% file: sections/2_related.tex
\section{Related Work}
\label{sec:relatedwork}

\noindent\textbf{Weakly-Supervised Semantic Segmentation.} 
Several works in the literature have proposed to use weak supervision to reduce the annotation cost. For semantic segmentation, one of the most popular forms are image-level labels, as they can be obtained with minimum human intervention. There are approaches that treat image-level labels with multiple instance learning techniques~\cite{pinheiro2015image, pathak2014fully, pathak2015constrained}, but these works achieve an accuracy far from their fully-supervised counterparts. Other works develop Expectation-Maximization methods to learn from weakly-annotated data~\cite{papandreou2015weakly}. More recently, a pool of works have focused on localizing class-specific cues with Class Activation Maps (CAMs)~\cite{zhou2016learning} in order to mine regions ~\cite{wei2017object, huang2018weakly, ahn2018learning, wei2018revisiting}, while others obtain regions with attention mechanisms ~\cite{zhang2018decoupled}.
Our model resembles to the work from \cite{wei2018revisiting}. Their pipeline consists of two networks, a deep neural network that produces pseudo-labels from CAMs, and a network that is trained with the obtained annotations. As our setup is semi-supervised, our first model will be trained with strong supervision only, while our second network will be trained with both pseudo- and strong annotations. 
For semantic segmentation, other weak signals have been exploited, such as scribbles~\cite{xu2015learning, lin2016scribblesup, tang2018normalized}, points~\cite{bearman2016s} or bounding boxes~\cite{papandreou2015weakly, dai2015boxsup, khoreva2017simple}. 

\noindent\textbf{Weakly-Supervised Instance Segmentation.} Few works have addressed weakly-supervised instance segmentation. Bounding box labels have been exploited by \cite{khoreva2017simple, zhao2018pseudo, li2018weakly} to recursively generate and refine pseudo-labels for the weak-labeled set. These methods typically rely on bottom-up segment proposals~\cite{pont2017multiscale, rother2004grabcut}. In contrast with this approach, \cite{remez2018learning} propose an adversarial scheme that learns to segment without using any object proposal technique. Although these works tackle weakly-supervised instance segmentation, their weak supervision consists in using bounding boxes, becoming the main challenge how to separate the foreground from the background within a bounding box. The only work that uses image-level supervision for weakly-supervised instance segmentation ~\cite{zhou2018weakly} detects peaks of CAMs and generates a query to retrieve the best candidate among a set of pre-computed object proposals (MCG)~\cite{pont2017multiscale}.

\noindent\textbf{Semi-Supervised Segmentation.} Semi-supervised learning allows to reduce the annotation burden while keeping a competitive performance. Some works that address weakly-supervised semantic segmentation present results for the semi-supervised case by combining their generated pseudo-annotations with a few strong labels~\cite{papandreou2015weakly, dai2015boxsup, khoreva2017simple, wei2018revisiting, li2018weakly}. Some other works exclusively tackle the semi-supervised scenario, as it is our case. Image-level labels were leveraged for semi-supervised semantic segmentation by \cite{hong2015decoupled}. Their pipeline consists of two separate networks, a classification and a segmentation network with bridged layers. They obtain remarkable results training with just a few strong annotations. The recent work
\cite{hu2017learning} proposes a new partially supervised training paradigm to combine bounding box annotations and pixel-level masks. To the authors knowledge, only \cite{hu2017learning, li2018weakly} have tackled semi-supervised instance segmentation. However, they assume a huge amount of weakly-labeled samples. In our work, we focus on low-budget scenarios, presenting the first results for semi-supervised instance segmentation for the Pascal VOC benchmark~\cite{everingham2010pascal} with no extra images from other datasets.

%% file: sections/3_benchmark.tex
\section{Benchmark for budget-aware segmentation}
\label{sec:benchmark}

The main focus of our work is to offer a unified analysis across different supervision setups and supervision signals for semantic and instance segmentation. Our motivation raises from the ultimate goal of weakly and semi-supervised techniques: the reduction of the annotation burden. 
We adopt the analysis framework from \cite{bearman2016s} and extend it to any supervision setup, applied to two different tasks: semantic and instance segmentation.

We estimate the annotation cost of an image from a well-known dataset for semantic and instance segmentation: the  Pascal VOC dataset~\cite{everingham2010pascal}. Our study considers four level of supervision: image-level, image-level labels + object counts, bounding boxes, and full supervision (i.e.~pixel-wise masks).
The estimated costs are inferred from three statistical figures about the Pascal VOC dataset drawn from~\cite{bearman2016s}: a) on average 1.5 class categories are present in each image, b) on average there are 2.8 objects per image, and c) there is a total of 20 class categories. 
Hence, the budgets needed for each level of supervision are:

\noindent\textbf{Image-Level (IL):} According to \cite{bearman2016s}, the time to verify the presence of a class in an image is of 1 second. The annotation cost per image is determined by the total number of possible class categories (20 in Pascal VOC). Then, the cost is of $ t_{IL}\, =\,20\,\mathrm{classes/image}\,\times\,1 \mathrm{s/class}\,=\,20\,\mathrm{s/image}$.


\noindent\textbf{Image-Level + Counts (IL+C):} IL annotations can be enriched by the amount of instances of each object class.
This scheme was proposed in for weakly-supervised object localization~\cite{gao2017c}, in which they estimate that the counting increases the annotation time to 1.48s per class.
Hence, the time to annotate an image with image labels and counts is $t_{IL+C} =\, t_{IL}\,+\,1.5\,\mathrm{classes/image}\,\times\,1.48 \, \mathrm{s/class}\,=\,22.22\,\mathrm{s/image}$.

\noindent\textbf{Full supervision (Full):} We consider the annotation time reported in 
~\cite{bearman2016s} for instance segmentation: 
$t_{Full} = 18.5 \, \mathrm{classes/image}\times 1 \mathrm{s/class} \, + \, 2.8\,\mathrm{mask/image} \, \times \, 79 \, \mathrm{s/mask} \, = \, 239.7 \, \mathrm{s/image}$. As we could not find any reference to the semantic segmentation task,  
we will assume that semantic segmentation labels require as much time as the 
instance segmentation ones.

\noindent\textbf{Bounding Boxes (BB):} Recent techniques have cut the cost of annotating a bounding box to 7.0 s/box by clicking the most extreme points of the objects~\cite{papadopoulos2017extreme}. Following the same reasoning as for dense predictions, the cost of annotating a Pascal VOC image with bounding boxes is $t_{bb} = 18.5 \, \mathrm{classes/image}\times 1 \mathrm{s/class} \, + \,2.8\,\mathrm{bb/image} \, \times\, 7 \, \mathrm{s/bb} \, = \, 38.1 \, \mathrm{s/image}$.

Table~\ref{tab:summary_costs} summarizes the average cost of the different supervision signals for a single Pascal VOC image. 

\begin{table}
\centering
\begin{tabular}{ccccc}
\toprule
    & \textbf{IL}  & \textbf{IL+C} & \textbf{Full}  & \textbf{BB}   \\
\bottomrule
Cost (s/image) & 20 & 22.22 &239.7  &38.1\\ 
\hline

\end{tabular}
\caption{Average annotation cost per image when using different types of supervision.}
\label{tab:summary_costs}
\end{table}

Given a certain annotation budget, the amount of annotated images will depend on the chosen level of supervision.
The lower the level of supervision, the more images will be annotated.
The central research question of our work is how to use an annotation budget: whether in few but fully supervised annotations, or in weaker labels for a larger amount of images. 

%% file: sections/4_semisupervised_setup.tex
\section{Semi-supervised segmentation}
\label{sec:setup}

Our pipeline consists of two different networks. A first fully supervised model $f_{\theta}$ is trained with strong-labeled samples from the ground truth $(X,Y) = \{(x_{1},y_{1}),...,(x_{N},y_{N})\}$, being $N$ the total number of strong samples. The network $f_{\theta}$ is an annotation network used to predict pseudo-labels $Y' = \{y'_{1}, ...,y'_{M}\}$ for $M$ unlabeled samples $X' = \{x'_{1}, ...,x'_{M}\}$. A second segmentation network $g_{\varphi}$ is trained with $(X,Y) \cup (X',Y')$,
as depicted in Figure~\ref{fig:scheme}. Depending on the task (semantic or instance segmentation), we will choose different architectures for the networks. It is important to remark that the proposed pipeline is independent to the network architecture used.

We present experiments for both the semantic and instance segmentation tasks for the Pascal VOC 2012 benchmark~\cite{everingham2010pascal}.
The standard semi-supervised setup adopted for this dataset consists of using the Pascal VOC 2012 train images (1464 images) as strong-labeled images, and an additional set (9118 images) from~\cite{hariharan2011semantic} as unlabeled/weak-labeled. In this section, we vary $N$ to analyze the performance at different annotation budgets, and consider $M$ to be the total size of the training dataset minus $N$~($M=10582-N$). Note that these $M$ samples are unlabeled, free of annotation cost.

%% file: sections/5_semanticseg.tex
\subsection{Semantic Segmentation}
\label{sec:sseg}

For semantic segmentation, we consider $f_{\theta}$ and $g_{\varphi}$ to have the same architecture, a DeepLab-v3+~\cite{chen2018encoder} with an Xception-65~\cite{chollet2017xception} encoder, with output stride of 16 for both training and evaluation.
We used the official TensorFlow implementation from ~\cite{chen2018encoder}. 
Following the setup described in Section~\ref{sec:setup}, we run experiments with the standard semi-supervised setup for Pascal VOC. Table~\ref{tab:sseg_deeplabv3} shows the results in terms of mean Intersection Over Union (mIoU) for different levels of supervision.
The first row sets the baseline of 
78.96 when training the annotation network $f_{\theta}$ (a DeepLab-v3+) with only the 1.4k images from the Pascal VOC 2012 train set. 
The next row, reports a mIoU of 79.41 when we train $g_{\varphi}$, also a DeepLab-v3+, with both the strong-labels $Y$ and the pseudo-labels $Y'$ obtained with $f_{\theta}$, which represents a small improvement. Finally, we trained a DeepLab-v3+ with all labels strongly-annotated (fully-supervised case), and obtained a mIoU of 80.42, close to the reference figure (81.21) reported in~\cite{chen2018encoder}.

\begin{table}[]
\centering
\resizebox{\linewidth}{!}{

\begin{tabular}{lcccc}
\toprule
                 & \textbf{\#Strong}     & \textbf{\#Unlabeled}     & \textbf{val mIoU}  & \textbf{test mIoU} \\ \hline
DeepLab-v3+ Ours & $\sim$1.4k &          & 
78,96   &77.26 \\
DeepLab-v3+ Ours & $\sim$1.4k & $\sim$9k & 79.41   &78.71\\
DeepLab-v3+ Ours & $\sim$10k  &          & 80.42   &80.29 \\ \hline
DeepLab-v3+~\cite{chen2018encoder}    & $\sim$10k  &     & 81.21 &- \\ \bottomrule
\end{tabular}
}
\caption{Performance of DeepLab-v3+ for the validation and test set of Pascal VOC 2012 with different supervision setups.}
\label{tab:sseg_deeplabv3}
\end{table}

\begin{figure}
\vspace*{-\baselineskip}
  \centering
  \includegraphics[width=\columnwidth]{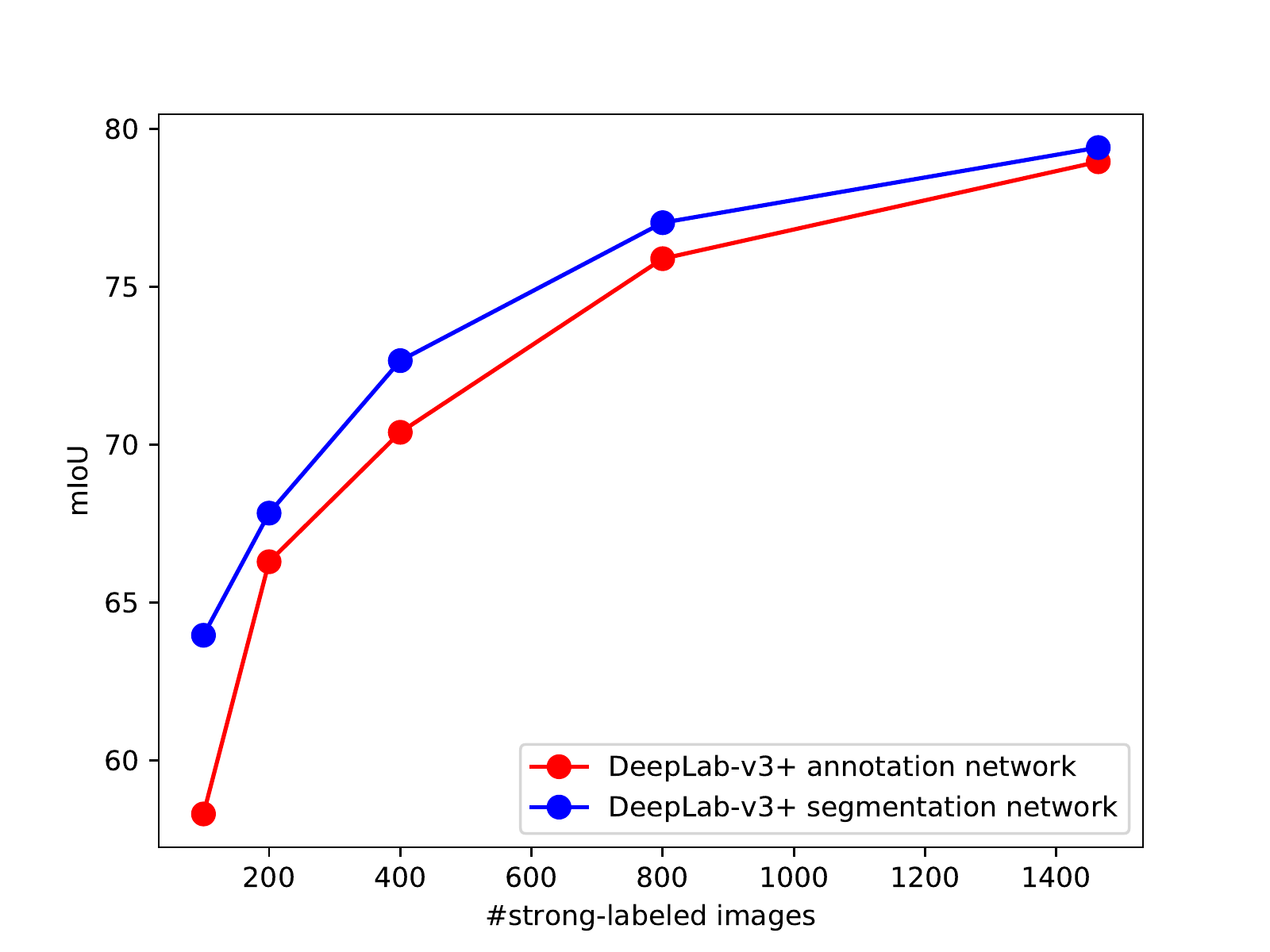}
  \caption{Semantic segmentation performance of the annotation and segmentation networks for an increasing budget for the validation set of Pascal VOC.}
  \label{fig:sseg}
\vspace*{-\baselineskip}
\end{figure}

To assess the impact of fixing different annotation budgets, we trained several DeepLab-v3+ $f_{\theta_N}$ with a varying number of strong-labeled training samples $N\in\{100,\,200,\,400,\,800,\,1464\}$. These networks are used to obtain pseudo-annotations for the $M$ samples without labels. 
In order to mitigate subset selection bias, for each value of $N$ we train 5 annotators with different random subsets of $N$ samples and report their average performance.
Then, we train a corresponding $g_{\varphi_N}$ for each $f_{\theta_N}$. Notice that the pseudo-annotations are obtained for free, as no supervision signal is required. 
Figure~\ref{fig:sseg} plots the obtained mIoU by the annotation network $f_{\theta_N}$ and segmentation network $g_{\varphi_N}$ for different annotation budgets.
We observe that, given a certain budget, the mIoU of $g_{\varphi_N}$ is always higher than the one obtained with the $f_{\theta_N}$ alone, and therefore the extra pseudo-labels improve the performance. 
This suggests that pseudo-annotations can increase the quality of the segmentation tool at no additional cost.


\begin{figure*}
\vspace*{-\baselineskip}
  \centering
  \includegraphics[width=\textwidth]{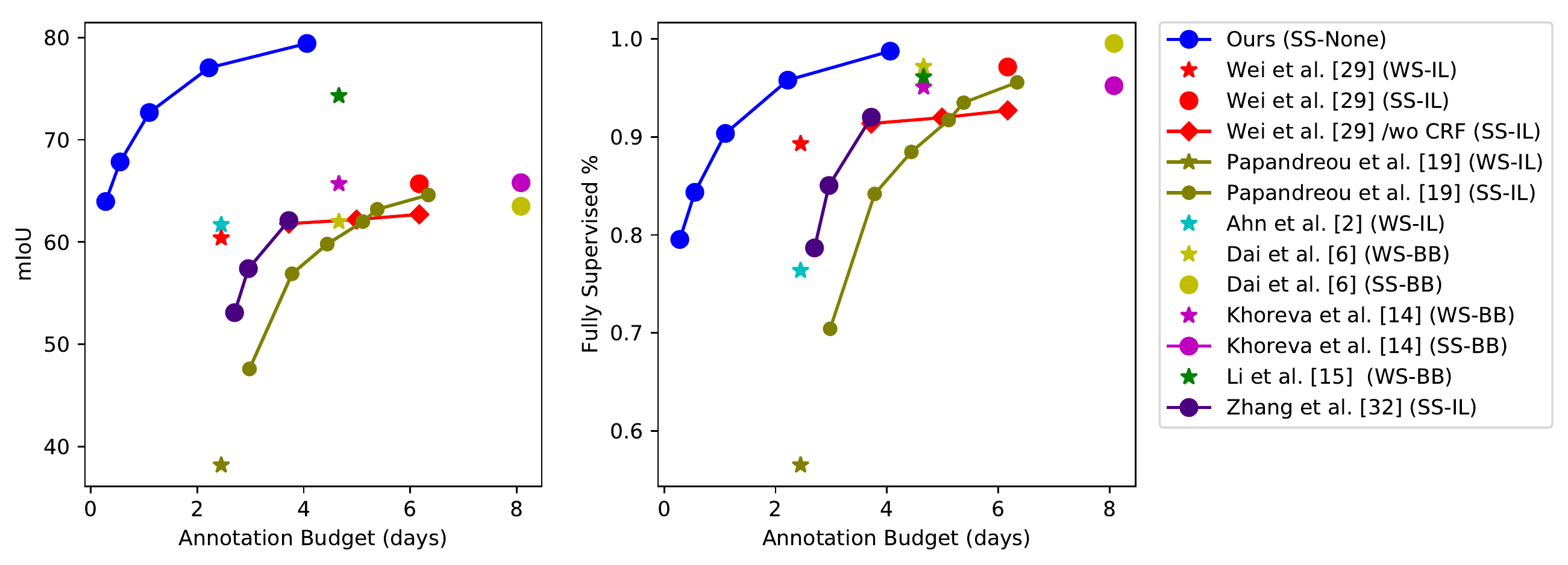}
  \caption{Semantic Segmentation comparison for the validation set of Pascal VOC with other semi-supervised (SS) and weakly-supervised (WS) methods, that use image-level labels (IL) or bounding box labels (BB).}
  \label{fig:sseg_comparison}
\vspace*{-\baselineskip}
\end{figure*}

\begin{figure}
  \centering
  \includegraphics[width=\columnwidth]{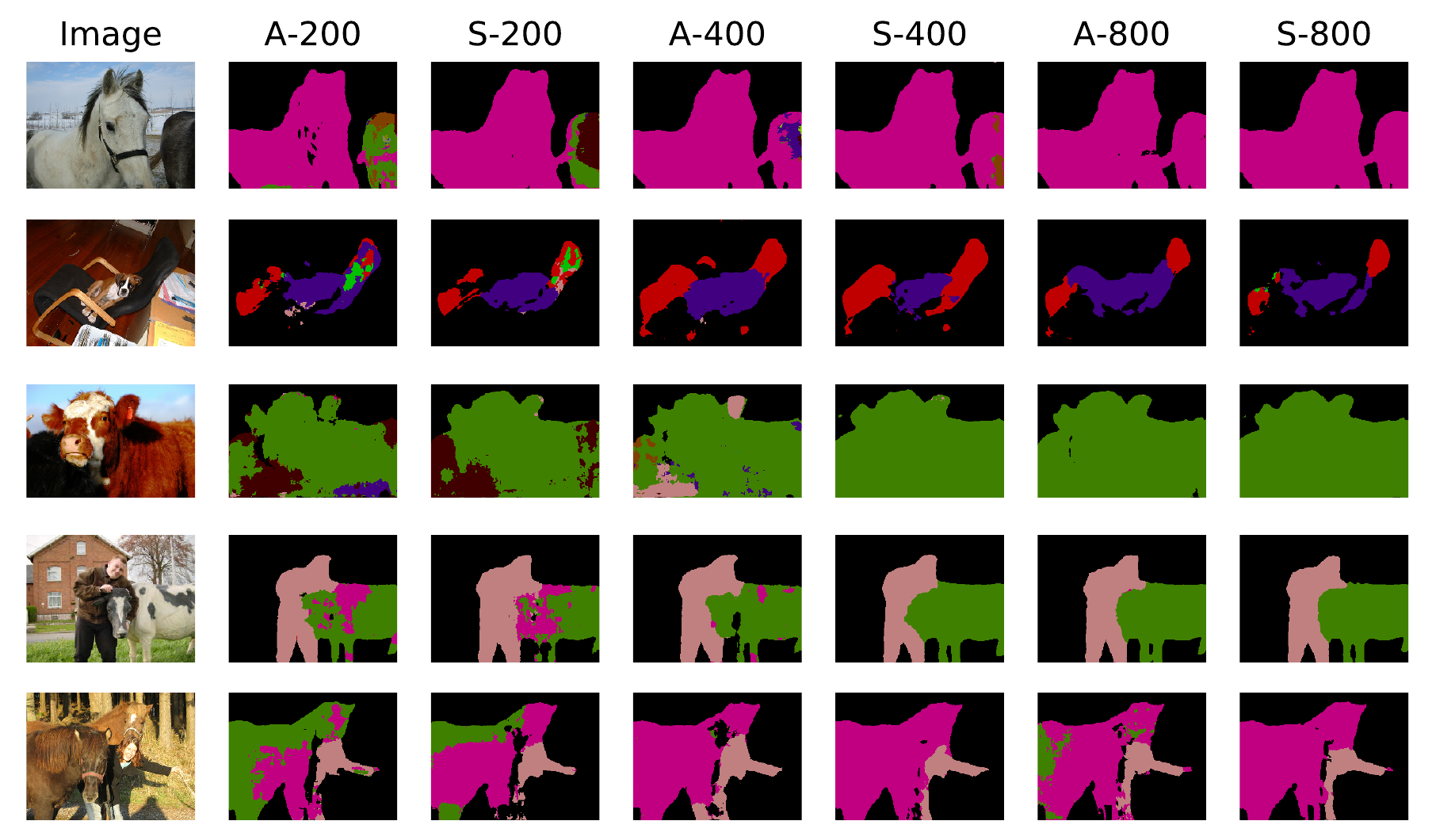}
  \caption{Visualization of Pascal VOC validation set for the annotation $f_{\theta_N}$ \textit{(A-)} and segmentation networks $g_{\varphi_N}$ \textit{(S-)}, depending on the number of strong labels used $N\in\{200,\,400,\,800$\} .}
  \label{fig:vis_sseg}
\vspace*{-\baselineskip}
\end{figure}

Figure~\ref{fig:sseg_comparison} compares our results with recent works of both weakly-supervised and semi-supervised approaches for semantic segmentation. The plot on the left shows the mIoU metric with respect to the annotation cost in days. We propose this analysis as a unified benchmark that allows a fair comparison between both weakly-supervised and semi-supervised pipelines.
We observe that our results obtained with DeepLab-v3+ outperform all previous methods (weakly or semi-supervised) at same or lower annotation budgets, setting a new state of the art of 79.41 mIoU for semi-supervised segmentation, using strong supervision only.
In order to compensate for the different network backbones used in the related works, Figure~\ref{fig:sseg_comparison} (right) normalizes the mIoU scores with the ones obtained by the fully-supervised counterparts. 
Our method with DeepLab-v3+ still reaches a closer number to the fully-supervised case compared to the other works at a fixed annotation budget. 
We want to highlight that our approach outperforms all weakly-supervised approaches when matching the annotation cost. Therefore, we conclude that it is preferable to invest the budget into collecting fewer fully supervised samples, than a larger amount of weakly-labeled ones. Figure~\ref{fig:vis_sseg} depicts some examples of semantic segmentation predicted by $f_{\theta_N}$ and $g_{\varphi_N}$ when using different number of strong labels.  As expected, $g_{\varphi_N}$ obtains better segmentation results than its counterpart $f_{\theta_N}$. We can also perceive that at low annotation budgets ($N=200$), the segments produced are able to accurately outline some contours, although the results are still far from the ones obtained with a higher $N$.

%% file: sections/6_instanceseg.tex
\subsection{Instance Segmentation}
\label{sec:ins_seg}

We will follow the semi-supervised pipeline described in Section \ref{sec:setup}, training an annotation network $f_{\theta}$ and a segmentation network $g_{\varphi}$.  This is the same scheme as in the semantic segmentation task presented in Section \ref{sec:sseg} but, in this case, we use the recurrent architecture for instance segmentation RSIS~\cite{salvador2017recurrent} for both $f_{\theta}$  and $g_{\varphi}$. We use the open-source code released by the authors. Further details about the RSIS architecture are presented in Section \ref{sec:semi_het}.

\begin{figure}
\centering
\includegraphics[width=\columnwidth]{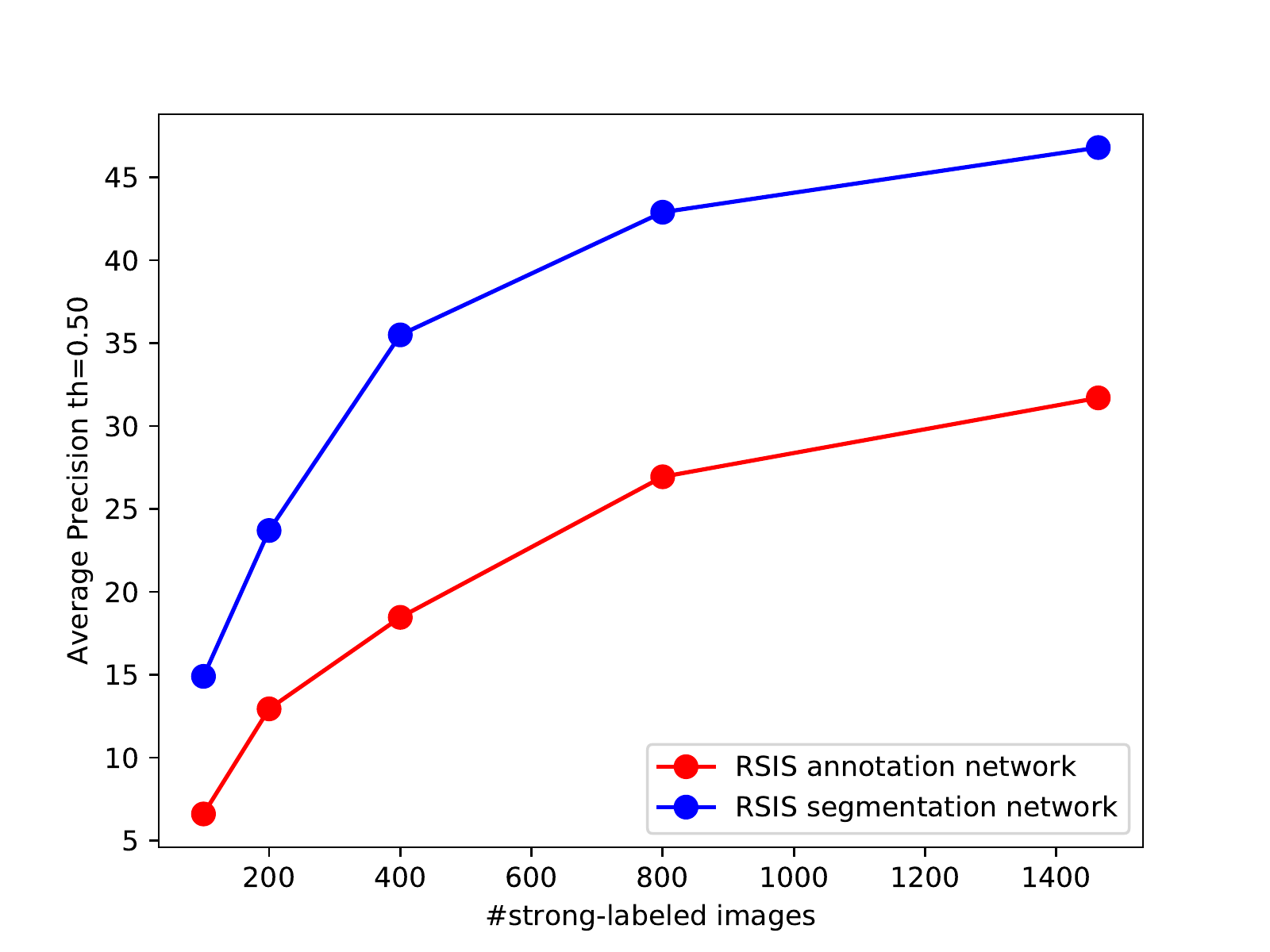}
\caption{Instance segmentation performance of the annotation and segmentation networks for an increasing budget for the validation set of Pascal VOC.}
\label{fig:iseg}
\vspace*{-\baselineskip}
\end{figure}

The results in Table \ref{tab:iseg_rsis_tab} show a similar behaviour to the semantic segmentation case from Table \ref{tab:sseg_deeplabv3}, although there is a more significant improvement of performance when the segmentation network $g_{\varphi_N}$ is trained with $(X,Y)\cup(X',Y')$, the union of the strong-labeled set and the pseudo-annotated set.
We follow the same setup presented in Section \ref{sec:sseg}, i.e. we consider different budget scenarios by varying the number of strong labels $N\in\{100,\,200,\,400,\,800,\,1464$\}. Figure~\ref{fig:iseg} reports the mean Average Precision at 0.5 for 5 different annotators trained with random splits of size $N$. The performance gap between $f_{\theta_N}$ and $g_{\varphi_N}$ is more significant for the instance segmentation task (Figure ~\ref{fig:iseg}) than for the semantic segmentation one (Figure \ref{fig:sseg}).
In the later, both curves converge when all available 1464 strong labels are used to train the annotation network, which indicates that the segmentation network does not learn anything new from the unlabeled images.
We hypothesize that learning instance segmentation is a more complex task, and more samples would be needed for both curves to converge.

\begin{table}[]
\centering
\begin{tabular}{lccc}
\toprule
                 & \textbf{\#Strong}     & \textbf{\#Unlabeled}     & \textbf{val AP 50}  \\ \hline
RSIS Ours & $\sim$1.4k &          & 
31.7\\
RSIS Ours & $\sim$1.4k & $\sim$9k & 
46.8\\
RSIS Ours & $\sim$10k  &          & 56.4 \\ \hline
RSIS~\cite{salvador2017recurrent}    & $\sim$10k  &          & 57.0 \\ \bottomrule
\end{tabular}
\caption{Performance of RSIS for the validation set of Pascal VOC 2012 with different supervision setups.}
\label{tab:iseg_rsis_tab}
\vspace*{-\baselineskip}
\end{table}

\begin{figure}
  \centering
  \includegraphics[width=\columnwidth]{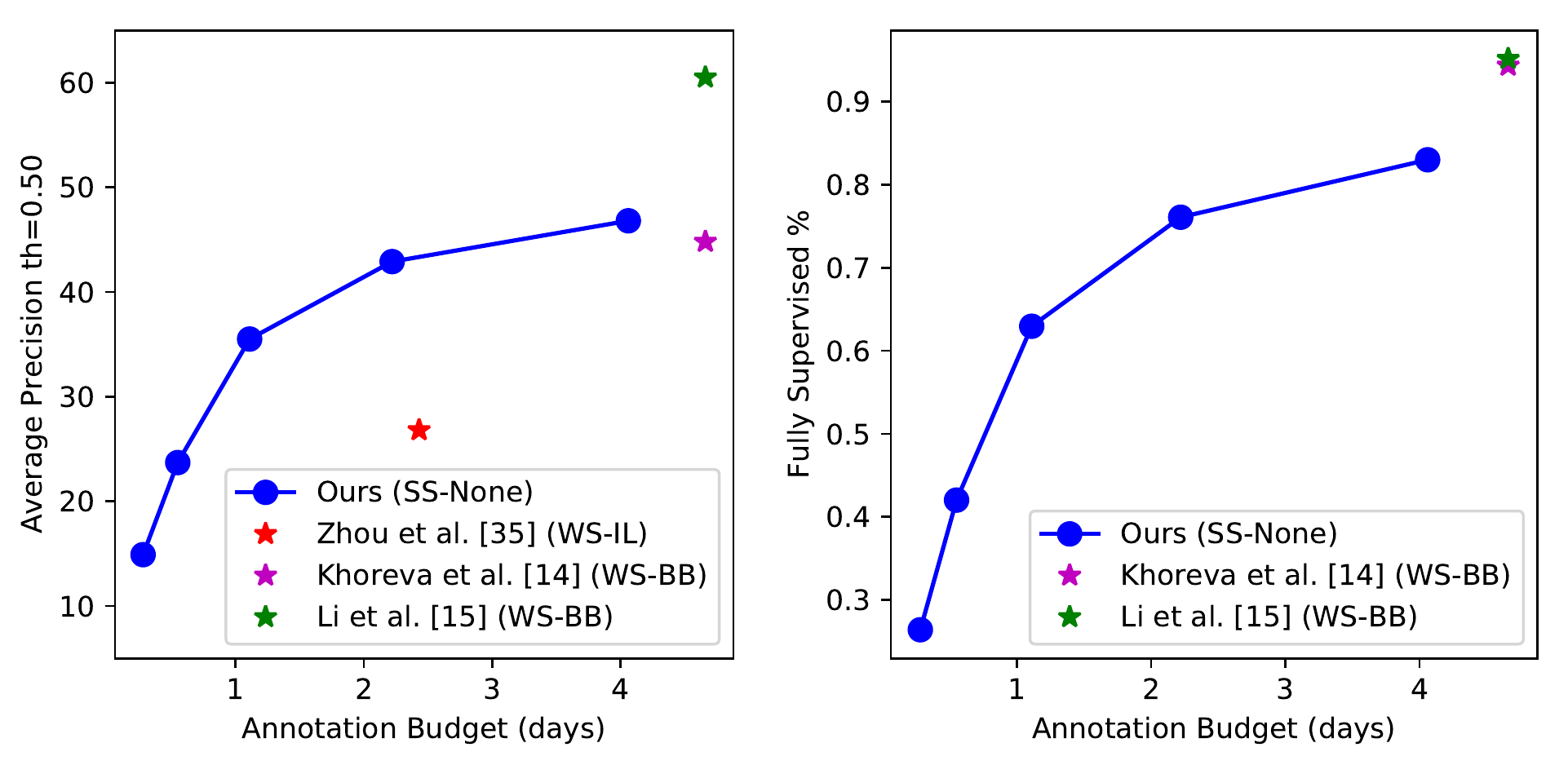}
  \caption{Instance Segmentation comparison for the validation set of Pascal VOC with other weakly-supervised (WS) methods, that use image-level labels (IL) or bounding box labels (BB).}
  \label{fig:iseg_comparison}
\end{figure}

\begin{figure}
  \centering
  \includegraphics[width=\columnwidth]{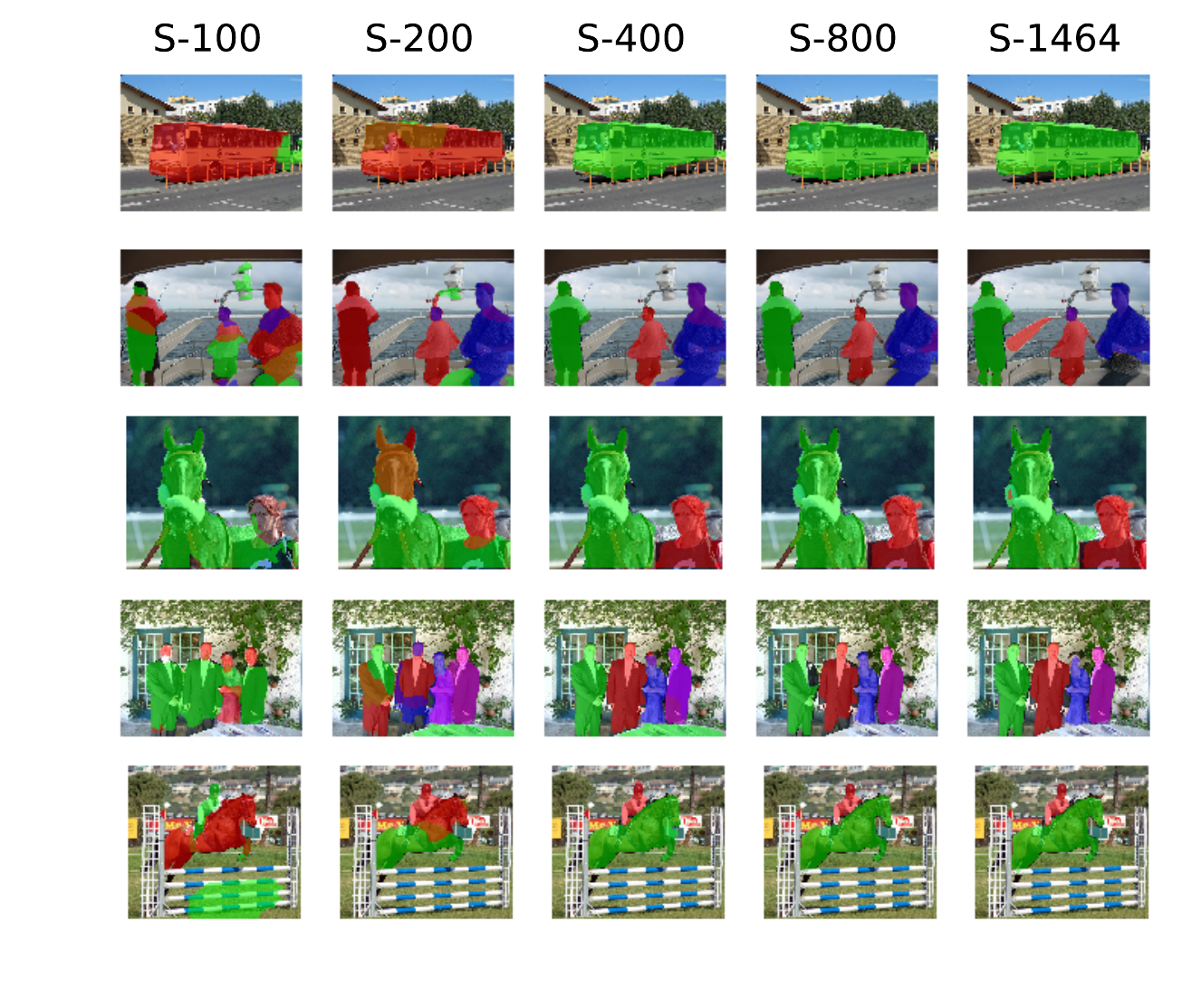}
  \caption{Visualization of Pascal VOC validation set for the instance segmentation network $g_{\varphi_N}$ \textit{(S-)} with $N\in\{100,\,200,\,400,\,800,\,1464$\} and $M=10582-N$. The AP (th=0.50) for each configuration is, from left to right, 
  of 14.9, 23.7, 35.7, 42.9 and 46.8.}
  \label{fig:vis_iseg}
\vspace*{-\baselineskip}
\end{figure}

Figure ~\ref{fig:iseg_comparison} compares our approach with related works that tackle weakly-supervised instance segmentation. For low annotation budgets there is the work from \cite{zhou2018weakly}, that addresses weakly-supervised instance segmentation with image-level labels. This task is clearly very challenging for the instance segmentation problem, and we demonstrate that when matching the annotation cost, our semi-supervised approach reaches significant better performance. We believe that working with a semi-supervised setup for low-annotation budgets is convenient for instance segmentation, as cheap labels such as image-level ones barely relate to distinguishing between different instances of an image. Bounding boxes, on the other hand, scale down the problem to separate the foreground from the background, but at the cost of more expensive annotations and thus at higher budgets ~\cite{khoreva2017simple, li2018weakly}. Figure ~\ref{fig:vis_iseg} depicts some examples predicted by the segmentation network $g_{\varphi_N}$ when varying $N$. The higher the $N$, the better the network distinguishes between different instances. 

\section{Training with heterogeneous annotations}
\label{sec:semi_het}

Heretofore, we have been assuming a semi-supervised setup where some samples are strongly-labeled and others are unlabeled. For instance segmentation, we observe in Figure \ref{fig:iseg} that the Average Precision for annotation networks $f_{\theta}$ trained with very few strong samples is very low (an annotation network trained with $N=100$ reaches a low figure of 
7.7 of AP). In this section we propose to use heterogeneous annotations, i.e., strong and weak annotations, instead of strongly-labeled samples alone. The main difference to our previous setup, is that now the annotation cost will come from two sources: the $N$ strong-labeled samples, and the $M$ weak-labeled ones. 

As weak labels, we choose image-level labels, in addition to knowing how many instances of each class category appear in an image (IL+C). This supervision signal was first employed for weakly-supervised object localization ~\cite{gao2017c}, and its annotation cost is almost the same as using simply image-level labels, as explained in Section~\ref{sec:benchmark}. To the best of our knowledge, this is the first time that image-level labels plus counts~(IL+C) is used as supervision for instance segmentation.

The setup is similar to the one explained in Section \ref{sec:setup}. For a better understanding, we will keep the same notation. Let $Z$ be the IL+C labels for the strongly-annotated subset $(X, Y, Z)$, and $Z'$ the IL+C labels for the weakly-annotated subset $(X', Z')$. To exploit the weak-labels $Z'$, now $f_{\theta}$ during training will receive as input $(X, Z)$, and will be optimized to predict $Y$. In order to infer the pseudo-annotations $Y'$, $(X', Z')$ will be fed into $f_{\theta}$. The segmentation network $g_{\varphi}$ works as in Section \ref{sec:setup}. The architecture of the annotation network $f_{\theta}$ is a modified version of RSIS~\cite{salvador2017recurrent}, and the architecture for the segmentation one corresponds to the original RSIS model. 

\begin{figure}
  \centering
  \includegraphics[width=\columnwidth]{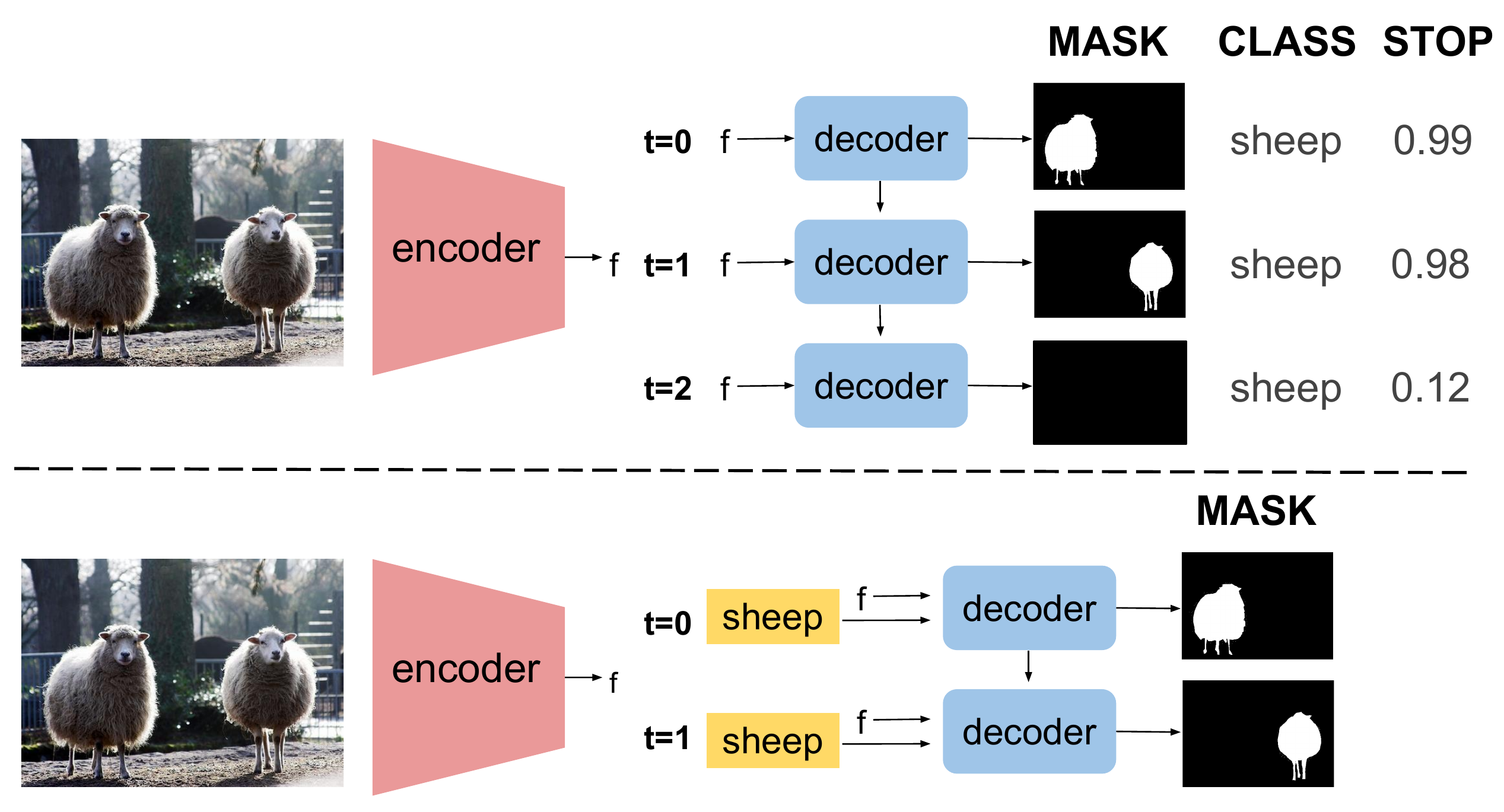}
  \caption{RSIS architecture in the first row, and W-RSIS architecture in the second. RSIS has three different outputs, the mask, class, and stop score. When the score is below a fixed threshold (e.g.~$\text{th}=0.5$), no more masks are produced. W-RSIS receives as input a token for each object in the image, so it only has the segmentation output.}
  \label{fig:wrsis-rsis}
\vspace*{-\baselineskip}
\end{figure}

\noindent\textbf{Annotation network.} RSIS ~\cite{salvador2017recurrent} consists in an encoder-decoder architecture~(Figure~\ref{fig:wrsis-rsis}). The encoder is a ResNet-101~\cite{he2016deep}, and the decoder is formed by a set of stacked ConvLSTM's~\cite{xingjian2015convolutional}. At each time step, a binary mask and a class category for each object of the image is predicted by the decoder. The architecture also has a stop branch that indicates if all objects have been covered. The main property of this architecture is that its output does not need any post-processing (as it happens with proposal-based methods, where proposals need to be filtered), so that the pseudo-annotation is the output of the network itself.
Our modified RSIS architecture for weak labels (W-RSIS) is also depicted in Figure~\ref{fig:wrsis-rsis}. The main difference is that, besides the features extracted by the encoder, the decoder receives at each time step a one-hot encoding of a class category representing each of the instances of the image. If there are several instances belonging to the same class, a one-hot encoding of that class will be given as input at several time-steps.

Table~\ref{table:abl}(a) presents an ablation study to analyze the impact of the different modifications included in W-RSIS.
We use the standard semi-supervised setup for Pascal VOC (1464 strong labels and 9118 weak labels).

The first row in Table~\ref{table:abl}(a) corresponds to the original RSIS, which annotates samples without using the weak labels. 
The \textit{+ IL} term means that the output of the \textit{softmax} class predictor is masked at inference time, thus constraining the possible classes predicted for the pseudo-labels. 
The option \textit{+ C} assumes that the count of instances \textit{n} in the image is known, and post-processes the pseudo-labels accordingly by keeping the first \textit{n} objects.
Finally, in W-RSIS the IL+C labels are an input of the network $f_{\theta}$, instead of simply being used as a post-processing step.
The ablation study shows how the proposed W-RSIS architecture maximizes the information contained in the IL+C weak labels. 

RSIS does not impose any order on the sequence of predicted masks. 

The permutation of the ground truth masks that leads to a lower loss with the predicted sequence is found with the Hungarian algorithm.

As in RSIS~\cite{salvador2017recurrent}, we
use the soft intersection over union loss (sIoU) as the cost
function between the mask predicted by our network 
and the ground truth mask.
Notice that now we have some restrictions in the sequence order, as we want an alignment between the input class category and the output, so in order to train W-RSIS, the Hungarian matching is performed only between ground truth instances of the same category.

Table~\ref{table:abl}(b) includes a second ablation study, in this case, about the masking of the Hungarian algorithm to just allow some permutations, constrained to class categories.

The first row corresponds to the basic case \textit{Hungarian}, but we observed that this did not constrain that our input classes were aligned with the classes of the predicted masks.

Afterwards, we applied the Hungarian algorithm among objects of the same category only, hence forcing an alignment between the input class categories and the actual class category of the prediction. 
This last \textit{Masked Hungarian} solution resulted to be the best option.

\begin{table}[]
    \begin{subtable}[b]{0.49\columnwidth}
    \centering
        \begin{tabular}[b]{lc}
        \toprule
                 & \textbf{AP 50} \\ 
        \midrule
        RSIS & 
        31.8  \\
        RSIS + IL  & 
        36.6\\
        RSIS + IL + C & 
        37.7\\
        W-RSIS  & 
        \textbf{40.0}\\ 
        \bottomrule
        \end{tabular}
        \caption{}
        \label{tab:abl_sub1}
    \end{subtable}
    \begin{subtable}[b]{0.49\columnwidth}
    \centering
        \begin{tabular}[b]{lc}
        \toprule
                          & \textbf{AP 50} \\ 
        \midrule
        Hungarian    & 
        35.2\\
        Masked Hungarian  & 
        \textbf{40.0}\\ 
        \bottomrule
        \end{tabular}
        \caption{}
        \label{tab:abl_sub2}
        \end{subtable}
\caption{(a) Ablation study of IL+C as inputs with the Pascal validation set. (b) Ablation study of different losses with the Pascal validation set. }
\label{table:abl}
\vspace*{-\baselineskip}
\end{table}

\begin{figure}
  \centering
  \includegraphics[width=\columnwidth]{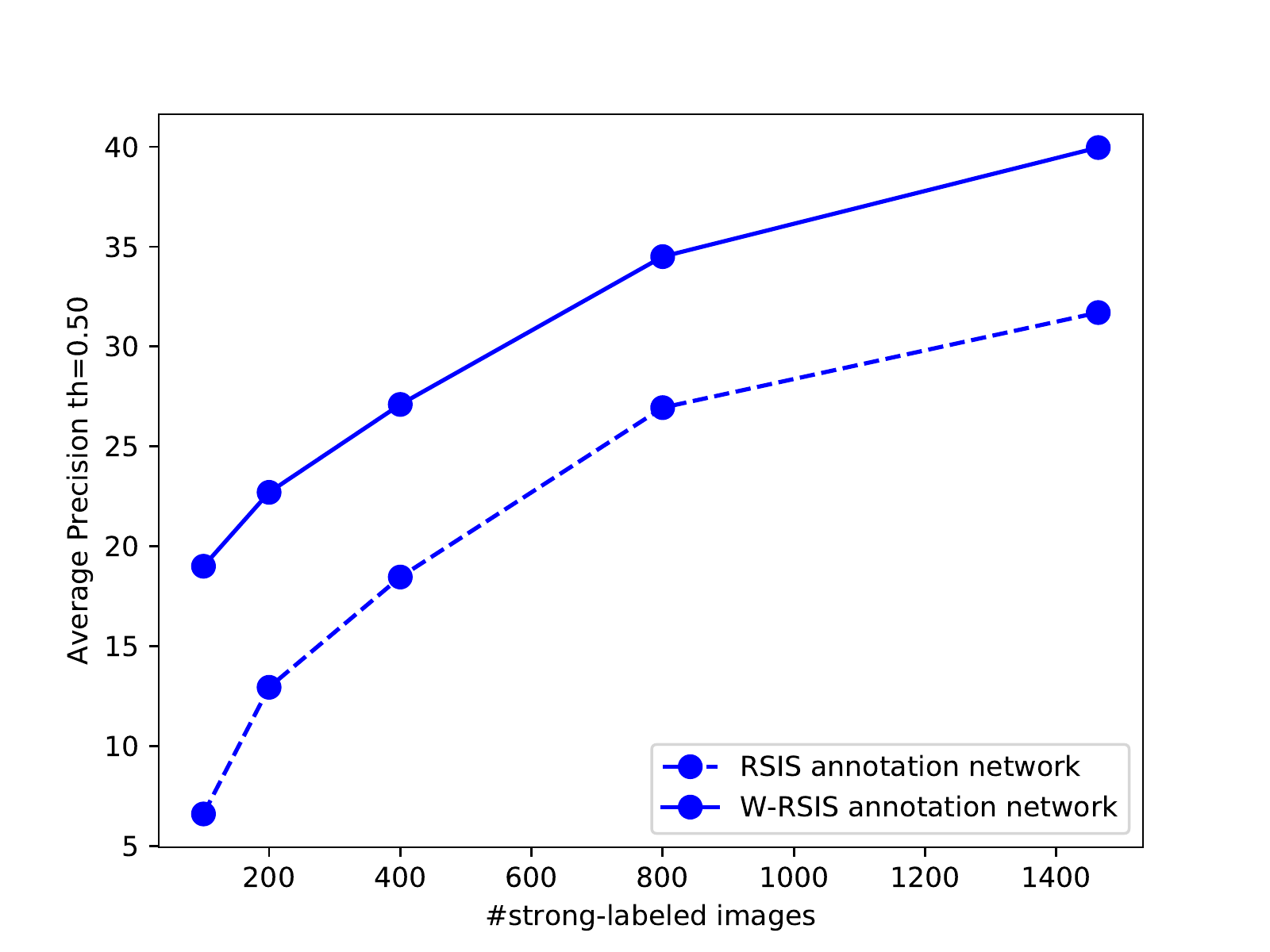}
  \caption{Comparison of RSIS annotation network, whose input are only the images to be annotated, and W-RSIS annotation network, whose input are the images and the IL+C information.}
  \label{fig:annotators_comparison}
\vspace*{-\baselineskip}
\end{figure}

Figure~\ref{fig:annotators_comparison} shows how W-RSIS generates better annotations compared to RSIS at different annotation budgets.
We train multiple W-RSIS models $f_{\theta_N}$ with a varying number of strong-labeled samples $N\in\{100,\,200,\,400,\,800,\,1464$\}, and compare them to the baseline RSIS. 
We notice that for any number $N$ of strong samples, W-RSIS outperforms RSIS. As in previous sections, for each $N$ we report the mean performance of 5 different models with different random subsets . 

\begin{figure}
  \centering
\includegraphics[width=\columnwidth]{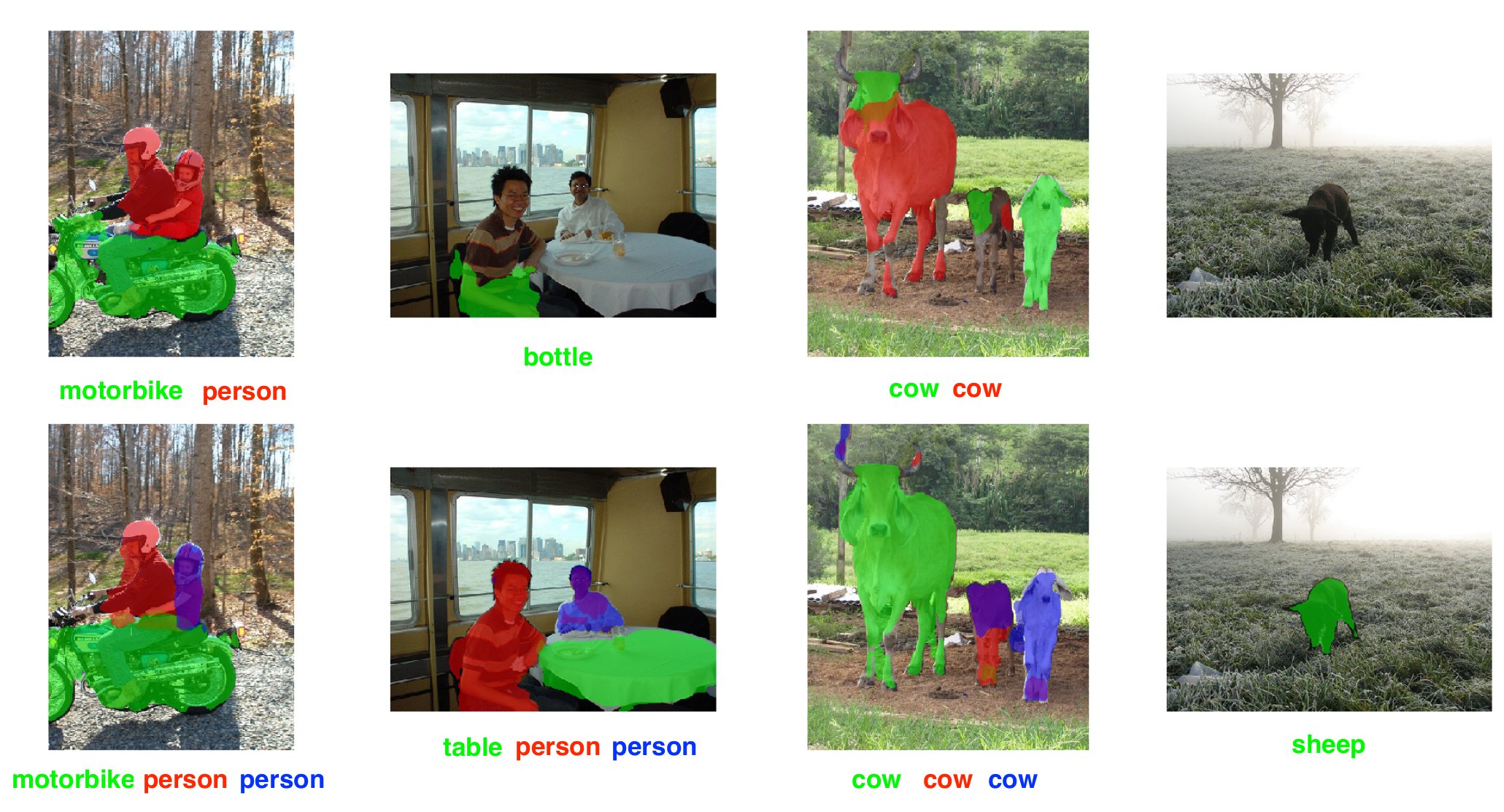}

    \caption{Comparison of pseudo-annotations obtained by RSIS (first row) and W-RSIS (second row) with $N=800$. The class category predicted for each pseudo-annotation is provided with the same color code.}
  \label{fig:rsis_wrsis}
\vspace*{-\baselineskip}
\end{figure}

Figure~\ref{fig:rsis_wrsis} shows qualitative results of the pseudo-annotations obtained by both configurations.  
The four pairs of images correspond to cases in which RSIS (first row) misses some of the instances because they were predicted with a low confidence score that does not reach the minimum detection threshold. 
W-RSIS (second row) does not present this limitation because the amount of instances for each class is provided by the weak annotation, so the confidence score is ignored.
Moreover, the second pair of images corresponds also to the case in which RSIS predicts a wrong class, a problem that W-RSIS does not have either as the category is already provided by the weak label.
The knowledge about the category of the pseudo-annotation provided by the class label facilitates the task, resulting in better quality masks. 


\noindent\textbf{Segmentation network.} We analyze the final performance of the segmentation network $g_{\varphi}$ in terms of the annotation cost when using the RSIS or W-RSIS annotation network. Notice that the segmentation network $g_{\varphi}$ is the same for both configurations (RSIS), only $f_{\theta}$ changes.

In Section \ref{sec:setup} annotating samples was cost-free, so varying the number $M$ did not impact the annotation budget. Consequently, we always considered $M$ to be the total size of the training set of Pascal VOC minus $N$~($M=10582 -N$). In the heterogeneous setup that we are considering now, there is a cost involved in annotating samples, as weak-labels are fed into the annotation network. 

In this section we will vary the number of weak-labeled samples $M\in\{912,\,2279,\,4459,\,6838,\,9118\}$, corresponding to the $\{10, 25, 50, 75, 100\} \%$ of the additional Pascal VOC set from ~\cite{hariharan2011semantic}.
In Table~\ref{tab:budget_comparisons} we fix different budget scenarios and we compare RSIS and W-RSIS.
We observe that for very low annotation budgets~($\sim$0,55 days), W-RSIS outperforms RSIS, meaning that it is more convenient to spend some budget on weak-labels and reduce the number of strong ones. We also notice that W-RSIS with $N=100$ strong samples and a few weak-labeled samples ($M=912$), reaches much higher results than those obtained with RSIS with $N=100$ and $M=10482$ unlabeled samples (25,2 vs. 14,9), but at a higher budget cost (0,51 vs. 0,27). For higher budgets the RSIS annotation network is strong enough, and does not benefit from weak labels.

\begin{figure}
  \centering

  \includegraphics[width=\columnwidth]{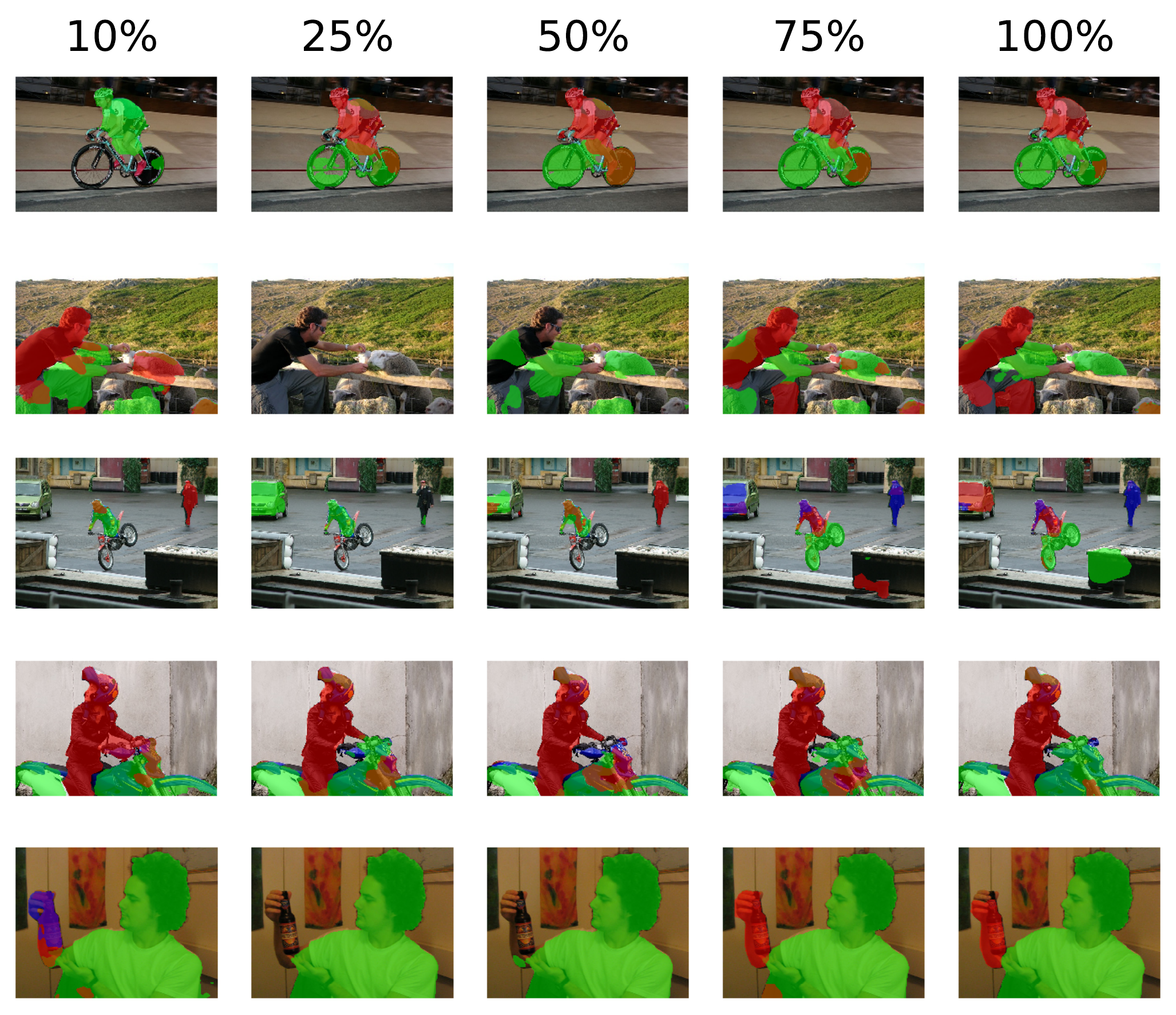}
   \caption{Visualization of Pascal VOC validation set for the instance segmentation network $g_{\varphi}$ when the annotation network is W-RSIS. The setup consists of $N=200$ and $M\in\{912,\,2279,\,4459,\,6838,\,9118\}$. The percentage indicates the fraction of $M$ compared to the total set~\cite{hariharan2011semantic}. The AP (th=0.50) for each configuration is, from left to right, of 
   27.3, 30.8, 30.7, 32.7 and 33.3.}
  \label{fig:vis_iseg_wrsis}
\vspace*{-\baselineskip}
\end{figure}

Table~\ref{tab:budget_comparisons} includes the results of ~\cite{zhou2018weakly}, the only previous work addressing the problem of instance segmentation with a low annotation budget. 
With both our models (RSIS or W-RSIS) we reach significant better results than ~\cite{zhou2018weakly}. At the same annotation budget ($\sim$2,2 days), RSIS with $N=200$ strong samples and $M=6838$ weak-labeled ones reaches a figure of 32,7 vs. 26,8 of AP 50. When having available more strong samples ($N=800$), RSIS reaches a higher figure of 42,9 vs. 26,8. Our models also outperform \cite{zhou2018weakly} at half its budget (35,5 or 30,8 vs. 26,8 of AP 50 at 1,1 vs. 2,43 annotation days). 
Figure~\ref{fig:vis_iseg_wrsis} shows qualitative results for W-RSIS for different numbers of weak-labeled samples.

\begin{table}[]
\centering
\resizebox{\linewidth}{!}{
\begin{tabular}{@{}lccccc@{}}
\toprule
                & \textbf{\#Strong} & \textbf{\#Unlabeled} & \textbf{\#Weak} & \textbf{Budget} & \textbf{AP 50} \\ \midrule
RSIS   & 100               & 10482                &-                 & 0.27            & 14.9           \\ \midrule
RSIS  & 200               & 10382                &-                 & 0.55            & 23.7           \\
W-RSIS & 100               &-                      & 912             & 0.51            & 25.2           \\ \midrule
RSIS   & 400               & 10182                &-                 & 1.1             & 35.5           \\
W-RSIS & 200               &-                      & 2279            & 1.14            & 30.8           \\ \midrule
RSIS   & 800               &9782                      &-             & 2.21            & 42.9           \\
W-RSIS & 200               &-                      & 6838            & 2.31            & 32.7           \\
Zhou et al.~\cite{zhou2018weakly}                &-                   &-                      & 10582           & 2.43            & 26.8           \\ \bottomrule
\end{tabular}
}
\caption{Results of the segmentation network when the annotation network changes (RSIS vs. W-RSIS) at different fixed annotation budgets (in days). 
}
\label{tab:budget_comparisons}
\end{table}

%% file: sections/7_conclusions.tex
\section{Conclusion}
\label{sec:conclusion}

The main contribution of this work is a unified benchmark for image segmentation structured around the annotation cost,
allowing to compare fairly weakly and semi-supervised methods. 

This budget-aware benchmark has allowed us to demonstrate that semi-supervised setups are preferable to weakly-supervised setups or, in other words, that fewer but strong labels achieve better results than a larger amount of weak labels.
This fewer labels paradigm is especially suitable in those domains in which collecting images is cumbersome (e.g. for the medical field). 

Moreover, the time to outline segments can be alleviated even further by modern interactive annotation tools ~\cite{acuna2018efficient, maninis2018deep}. Therefore, at a restricted annotation cost, more strong labels can be obtained, aiming at closer figures compared to the fully-supervised case.

\noindent \textbf{Acknowledgments:} This research was supported by the Spanish Ministry of Economy and Competitiveness and the European Regional Development Fund (TEC2016-75976-R, TIN2015-65316-P), by the BSC-CNS Severo Ochoa program SEV-2015-0493, by LaCaixa-Severo Ochoa International Doctoral Fellowship program, and by the Catalan Government (2017-SGR-1414).